\documentclass{article}
\usepackage{spconf,amsmath,graphicx}
\usepackage{algorithm2e}
\usepackage{hyperref}
\usepackage{verbatim}
\usepackage{subcaption}
\usepackage{amssymb}
\usepackage{multirow}
\usepackage{xcolor}
\usepackage{enumitem}
\usepackage{dcolumn}
\usepackage{threeparttable}
\usepackage{booktabs}
\usepackage{url,footnote}
\usepackage{breqn}
\usepackage{here}

\title{A CROSS-MODAL FUSION NETWORK BASED ON SELF-ATTENTION AND RESIDUAL STRUCTURE FOR MULTIMODAL EMOTION RECOGNITION}
\name{
    Ziwang Fu$^{1}$, Feng Liu$^{2, 3, 4, *}$\thanks{* Corresponding author.}, Hanyang Wang$^{2}$, Jiayin Qi$^{1, 3, *}$, Xiangling Fu$^{1, *}$, Aimin Zhou$^{2, 4, *}$, Zhibin Li$^{2}$
    \vspace{-5pt}
}

\address{
    $^1$ Beijing University of Posts and Telecommunications, Beijing, China\\
    $^2$ East China Normal University, Shanghai, China\\
    $^3$ Shanghai University of International Business and Economics, Shanghai, China\\
    $^4$ Shanghai Key Laboratory of Mental Health and Psychological Crisis Intervention, School of \\ Psychology and Cognitive Science, East China Normal University, Shanghai, China
    \vspace{-15pt}
}
\begin{document}
\ninept
\maketitle
\begin{abstract}

The audio-video based multimodal emotion recognition has attracted a lot of attention due to its robust performance. Most of the existing methods focus on proposing different cross-modal fusion strategies. However, these strategies introduce redundancy in the features of different modalities without fully considering the complementary properties between modal information, and these approaches do not guarantee the non-loss of original semantic information during intra- and inter-modal interactions. In this paper, we propose a novel cross-modal fusion network based on self-attention and residual structure (CFN-SR) for multimodal emotion recognition. Firstly, we perform representation learning for audio and video modalities to obtain the semantic features of the two modalities by efficient ResNeXt and 1D CNN, respectively. Secondly, we feed the features of the two modalities into the cross-modal blocks separately to ensure efficient complementarity and completeness of information through the self-attention mechanism and residual structure. Finally, we obtain the output of emotions by splicing the obtained fused representation with the original representation. To verify the effectiveness of the proposed method, we conduct experiments on the RAVDESS dataset. The experimental results show that the proposed CFN-SR achieves the state-of-the-art and obtains 75.76\% accuracy with 26.30M parameters. Our code is available at \url{https://github.com/skeletonNN/CFN-SR}.

\end{abstract}
\section{Introduction}
\label{sec:intro}

Multimodal emotion recognition has attracted a lot of attention due to its high performance and robustness \cite{dai2021multimodal}, which is applied in various fields such as human-computer interaction and social robotics \cite{zhang2019deep}. The main goal of multimodal emotion recognition is to obtain human emotion expressions from a video sequence. Humans express their emotions mainly through multiple modalities such as speech \cite{cao2021hierarchical}, body gestures \cite{wu2020generalized}, facial expressions \cite{hernandez2021multi}, and text \cite{naseem2020transformer}. Although many studies have employed more complex modalities, video and audio are still the primary modalities used for this task due to their ability to adequately convey emotion. Therefore, in this work, we focus on the audio-video based multimodal emotion recognition.

Among the existing researches, multimodal emotion recognition can be classified according to the modal information fusion method: early fusion \cite{tripathi2018multi}, late fusion \cite{zhang2017learning, atmaja2020multitask} and model fusion \cite{liu2021multimodal, sun2021multimodal}. Early fusion is to extract and construct multiple modal data into corresponding modal features before stitching them together into a feature set that integrates each modal feature. Late fusion is to find out the plausibility of each model, and then to coordinate and make joint decisions. Recently, with the development of Transformer \cite{10.5555/3295222.3295349} for natural language processing and computer vision tasks, model fusion is often done using Transformer for cross-modal interactions, with significant performance improvements due to the flexibility of fusion locations. For audio-video emotion recognition, \cite{huang2020multimodal} introduced Transformer to fuse audio and video representations. Different cross-modal fusion strategies were explored by \cite{zhou2019exploring}. \cite{liu2021multimodal} proposed a novel representation fusion method, Capsule Graph Convolutional Network(CapsGCN), to use graph capsule networks for audio and video emotion recognition.

However, existing methods ignore the complementary information between different modalities, and the final decision often requires the joint decision of the two modality-specific features as well as the fused features. The complementary information ensures that there is an improvement in performance when the fusion module is added. In addition, there is still a significant amount of redundancy in the way existing methods fuse. Some models are only stitched together for the final result output, and the stitched result will have many duplicate representations, and the feature information needs to be filtered to further reduce the redundant features before stitching. Moreover, the existing methods cannot guarantee the information integrity in the feature learning process, and often the learning of intra- and inter-modal information will lose some semantic information.

\begin{figure*}[!t]
\centering
\vspace{-0.5em}
\includegraphics[scale =0.6]{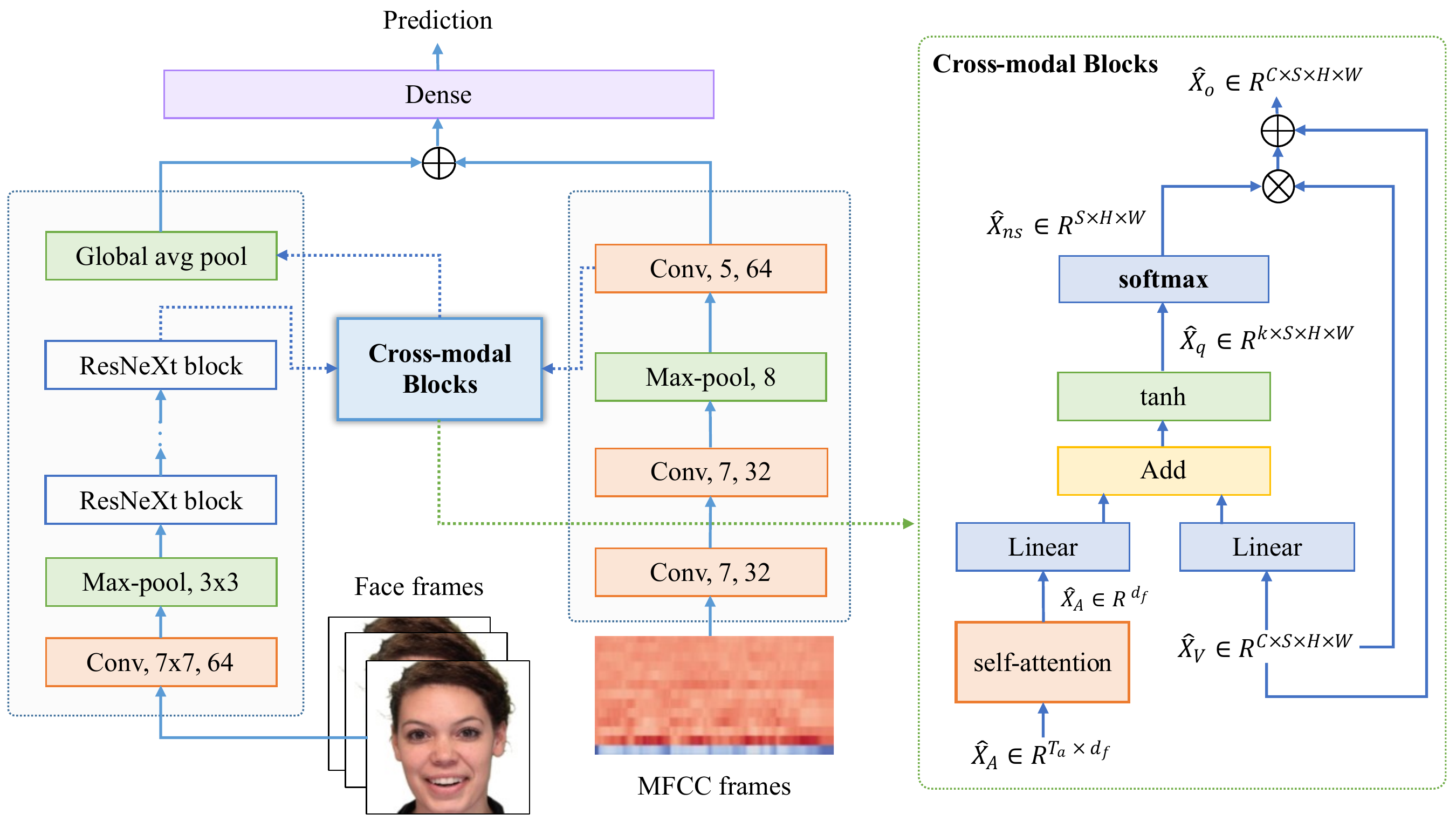}
\caption{The overall architecture of CFN-SR. \textbf{Left}: the flow structure of the whole model, extracting the higher-order semantic features of video and audio by ResNeXt and 1D CNN, respectively. \textbf{Right}: the cross-modal fusion blocks, which enables the complementarity and completeness of modal interactions to play a role through the introduction of self-attention mechanism and residual structure.}
\vspace{-0.5em}
\label{fig:framework}
\end{figure*}

To solve the above problems, we propose a novel cross-modal fusion network based on self-attention and residual structure (CFN-SR) for multimodal emotion recognition. Specifically, we first perform representation learning for audio modality and video modality. The spatio-temporal structure features of video frame sequences are obtained by ResNeXt \cite{xie2017aggregated}. The MFCC features of audio sequences are obtained by a simple and effective 1D CNN. Secondly, we feed the features into the cross-modal blocks separately, and make the audio modality perform intra-modal feature selection through a self-attention mechanism, which will enable the selected features to interact with the video modality adaptively and efficiently between modalities. The integrity of the original structural features of the video modality can be ensured by the residual structure. Finally, we obtain the output of emotions by splicing the obtained fused representation and the original representation.We apply the model to the RAVDESS \cite{livingstone2018ryerson} multimodal emotion recognition dataset, and the experimental results demonstrate that our proposed CFN-SR is more effective, and our method achieves state-of-the-art compared with other models, obtaining an accuracy of 75.76\% with a parametric number of 26.30M.

\section{Methodology}
\label{sec:format}

As shown in Figure \ref{fig:framework}, we design a cross-modal fusion network based on self-attention and residual structure. Firstly, we use the 3D CNN to obtain the video features and the 1D CNN to obtain the audio features. Then, we obtain the inter- and intra-modal fusion representations of the two modalities by the cross-modal fusion blocks. Finally, we obtain the output of the emotions by splicing the obtained fusion representation with the original representation. We will describe this process in detail below.

\subsection{Audio Encoder}

For the audio modality, recent work \cite{neverova2015moddrop, wang2020speech} has demonstrated the effectiveness of deep learning methods based on Mel Frequency Cepstrum Coefficient (MFCC) features. We design a simple and efficient 1D CNN to perform MFCC feature extraction. Specifically, we use the feature preprocessed audio modal features as input, denoted as $X_{A}$. We first pass the features through a 2-layer convolution operation to extract the local features of adjacent audio elements. After that, we use the max-pooling to downsample, compress the features, and remove redundant information. The specific equation is as follows:

\begin{equation}
    \hat{X}_{A} = BN(ReLU(Conv1D(X_{A}, k_{A})))
\end{equation}
\begin{equation}
    \hat{X}_{A} = Dropout(BN(MaxPool (\hat{X}_{A})))
\end{equation}
where $BN$ stands for Batch Normalization, $k_{A}$ is the size of the convolution kernel of modality audio and $\hat{X}_{A}$ denotes the learned semantic features. We again feed the learned features into a 1D temporal convolution to obtain the higher-order semantic features of the audio. Finally, we flatten the obtained features:

\begin{equation}
    \hat{X}_{A} = Flatten(BN(ReLU(Conv1D(\hat{X}_{A}, k_{A})))))
\end{equation}

\subsection{Video Encoder}

Video data are dependent in both spatial and temporal dimensions, thus a network with 3D convolutional kernels is needed to learn facial expressions and actions. We consider both the performance and training efficiency of the network and choose the 3D ResNeXt \cite{xie2017aggregated} network to obtain the spatio-temporal structural features of video modalities. ResNeXt proposes a group convolution strategy between the deep segmentation convolution of ordinary convolutional kernels, and achieves a balance between the two strategies by controlling the number of groups with a simple structure but powerful performance. We use feature preprocessed audio modal features as input, denoted as $X_{V}$. We obtain the higher-order semantic features of video modalities by this network:

\begin{equation}
    \hat{X}_{V} = ResNeXt50(X_{V}) \in \mathbb{R}^{C \times S \times H \times W}
\end{equation}
where $\hat{X}_{V}$ denotes the learned semantic features, $C$, $S$, $H$ and $W$ are the number of channels, sequence length, height, and width, respectively. After obtaining the higher-order semantic features, we feed them into the cross-modal blocks and fuse the audio feature representations. We believe that the final fused representation obtained contains not only the higher-order semantic features of the video modality, but also the interactive features of the two modalities. After that, we perform downsampling using the average pool to reduce redundant information.

\subsection{Cross-modal Blocks}
\label{sec: cs}

Through encoding operations for both modalities, we obtain higher-order semantic features for both audio and video modalities. To be able to make the final decision more accurate, we exploit the complementary intra- and inter-modal interaction information between the two modalities. Specifically, we first make the audio modality undergo intramodal representation learning through a self-attention mechanism. This operation allows adaptive learning of higher-order semantic features of the audio modality, making it more focused on features that have a greater impact on the outcome weights. Following Transformer \cite{10.5555/3295222.3295349}, we use self-attention to perform feature selection on audio features. The self-attention mechanism is able to reflect the influence of feature neighboring elements, whose query, key, and value are all representations of the same audio modality under different projection spaces. The specific formula is as follows:

\begin{equation}
    self-attention(Q,K,V) = softmax(\frac{QK^T}{\sqrt{d_k}})V
\end{equation}
where $Q, K, V$ denotes $Z^{\left[ i-1 \right]}_{A}$. $Z^{\left[ i-1 \right]}_{A}$ is represented by different projection spaces with different parameter matrices, where $i$ represents the number of layers of transformer attention, $i=1,...,D$. We feed the learned weight information into the full connection to obtain the feature adaptive learning results:

\begin{equation}
    \hat{X}_{A} = LN( Z^{\left[ i \right]}_{A} + Feedforward(Z^{\left[ i \right]}_{A})) \in \mathbb{R}^{d_f}
\end{equation}
where $LN$ represents layer normalization, $d_f$ represents dimensions of the extracted features. This operation enables the higher-order features of the audio modality to perform feature selection, making it more focused on features that have a greater impact on the outcome.

Then, we make the automatically selected features and the video modality perform efficient inter-modal interactions. The module accepts input for two modalities, which is called $\hat{X}_A \in \mathbb{R}^{d_f}$ and $\hat{X}_V \in \mathbb{R}^{C \times S \times H \times W}$. We obtain the mapping representations of the features for the two modalities by a linear projection. And then we process the two representations by $add$ and $tanh$ activation function. Finally, the fused representation $\hat{X}_o \in \mathbb{R}^{C \times S \times H \times W}$ is obtained through $softmax$. We believe that the final fused information contains not only the complementary information of the two modalities, but also the features of the video modality. The specific formula is as follows:

\begin{equation}
    \hat{X}_q = tanh((W_v\hat{X}_V + b_v) + W_a\hat{X}_A) \in \mathbb{R}^{k \times S \times H \times W}
\end{equation}
\begin{equation}
    \hat{X}_o = (softmax(\hat{X}_q) \otimes \hat{X}_V) \oplus \hat{X}_V \in \mathbb{R}^{C \times S \times H \times W}
\end{equation}
in which $W_v \in \mathbb{R}^{k \times C}$ and $W_a \in \mathbb{R}^{k \times d_f}$ are linear transformation weights, and $b_v \in \mathbb{R}^{k}$ is the bias, where $k$ is a pre-defined hyper-parameter, and $\oplus$ represents the broadcast addition operation of a tensor and a vector.

This operation enables high-level features to learn complementarily with low-level features, which complement the semantic features of low-level features and express richer information. In this process, to ensure that the information of the video modality is not lost, we ensure the integrity of the original structural features of the video modality through the residual structure.

\subsection{Classification}

Finally, we obtain the output of emotions $I = [\hat{X}_o, \hat{X}_A]$ by splicing the obtained fused representation and the original representation. The cross-entropy loss is used to optimize the model. The specific equation is shown as follows:

\begin{equation}
    prediction = W_1I + b_1 \in {R}^{d_{out}}
\end{equation}
\begin{equation}
    L = -\sum_{i} y_ilog(\hat{y}_i)
\end{equation}
where $d_{out}$ is the output dimensions of emotional categories, $ W_1 \in {R}^{d_{out}}$ is the weight vectors, $b_1$ is the bias, $y = \{y_1, y_2, ..., y_n\}^T$ is the one-hot vector of the emotion label, $\hat{y} = \{\hat{y}_1, \hat{y}_2, ..., \hat{y}_n\}^T$ is the predicted probability distribution, $n$ is the number of emotion categories.

\section{EXPERIMENTS}

\subsection{Datasets}

The Ryerson Audio-Visual Database of Emotional Speech and Song (RAVDESS) \cite{livingstone2018ryerson} is a multimodal emotion recognition dataset containing 24 actors (12 male, 12 female) of 1440 video clips of short speeches. The dataset is performed when the actors are told the emotion to be expressed, with high quality in terms of both video and audio recordings. Eight emotions are included in the dataset: neutral, calm, happy, sad, angry, fearful, disgust and surprised.

\subsection{Implementation Details}

For the video modality, we extract 30 consecutive images from each video. We crop the face region using the 2D face markers provided for each image and then resize to (224,224). Data augmentation is performed using random cropping, level flipping and normalization methods. For the audio modality, since the first 0.5 seconds usually do not contain sound, we trim the first 0.5 seconds and keep it consistent for the next 2.45 seconds. Following the suggestion of \cite{jin2015speech}, we extract the first 13 MFCC features for each cropped audio clip.

We perform 5-fold cross-validation on the RAVDESS dataset to provide more robust results. We divide the 24 actors into a training and a test set in a 5:1 ratio. Since the actors' gender is represented by an even or odd number of actor IDs, we enable gender to be evenly distributed by rotating 4 consecutive actor IDs as the test set for each fold of cross-validation. The model is trained using the Adam optimizer \cite{kingma2014adam} with a learning rate of 0.001, and the entire training of the model is done on a single NVIDIA RTX 8000. The final accuracy reported is the average accuracy over the 5 folds.

\subsection{Baselines}

For this task, we implement multiple recent multimodal fusion algorithms as our baselines. We categorize them into the following:
\begin{enumerate}
    \item Simple feature concatenation followed by fully connected layers based on \cite{ortega2019multimodal} and MCBP \cite{fukui2016multimodal} are two typical early fusion methods.
    \item Averaging and multiplication are the two standard late fusion methods that are adopted as the baselines.
    \item Multiplicative layer \cite{liu2018learn} is a late fusion method that adds a down-weighting factor to CE loss to suppress weaker modalities.
    \item MMTM \cite{joze2020mmtm} module allows slow fusion of information between modalities by adding to different feature layers, which allows the fusion of features in convolutional layers of different spatial dimensions.
    \item MSAF \cite{su2020msaf} module splits each channel into equal blocks of features in the channel direction and creates a joint representation that is used to generate soft notes for each channel across the feature blocks.
    \item ERANNs \cite{verbitskiy2021eranns} is the state-of-the-art method by proposing a new convolutional neural network architecture for audio-video emotion recognition.
\end{enumerate}

\subsection{Comparison to State-of-the-art Methods}

Table \ref{tab:avg} shows the accuracy comparison of the proposed method with baselines. From the table \ref{tab:avg}, we can see that our model achieves an accuracy of 75.76\% with 26.30M number of parameters, reaching the state-of-the-art. Compared to the unimodal baseline, our network accuracy exceeds more than 10\%, verifying the importance of multimodal fusion. In addition, the incorporation of the cross-modal fusion block only introduces a 30K number of parameters, but brings a significant performance improvement. Compared with early fusion methods, our method has more than 4\% accuracy improvement, which shows that finding the association between video and audio modalities is a difficult task in the early stage. What's more, our proposed CFN-SR has 2.64\% improvement over MMTM \cite{joze2020mmtm} and 0.9\% improvement over the best-performing MSFA \cite{su2020msaf}, and both have comparable number of parameters. This indicates that our model fully exploits the complementarity between the two modalities and reduces the redundant features.

\begin{table}[t]
    \fontsize{9}{11}\selectfont
    \centering
    \caption{\fontsize{9}{11}\selectfont Comparison between multimodal fusion baselines and
ours for emotion recognition on RAVDESS.
    }
    \label{tab:avg}
    \begin{tabular}{lcccc}
        \toprule
        Model & Fusion Stage & Accuracy & \#Params \\ 
        \midrule
        3D RexNeXt50 (Vid.) & - & 62.99 & 25.88M \\
        1D CNN (Aud.) & - & 56.53 & 0.03M \\
        \hline
        Averaging & Late & 68.82 & 25.92M \\
        Multiplicative $\beta$=0.3 & Late & 70.35 & 25.92M \\
        Multiplication & Late & 70.56 & 25.92M \\
        Concat + FC \cite{ortega2019multimodal} & Early & 71.04 & 26.87M \\
        MCBP \cite{fukui2016multimodal} & Early & 71.32 & 51.03M \\
        MMTM \cite{joze2020mmtm} & Model & 73.12 & 31.97M \\
        MSAF \cite{su2020msaf} & Model & 74.86 & 25.94M \\
        ERANNs \cite{verbitskiy2021eranns} & Model & 74.80 & - \\
        \hline
        \textbf{CFN-SR  (Ours)} & Model & \textbf{75.76} & \textbf{26.30M} \\
        \bottomrule
    \end{tabular}%
\end{table}

\subsection{Ablation Study}

Table \ref{tab:pool} shows the ablation experiments on the RAVDESS dataset. To verify the effectiveness of the cross-modal blocks, we obtain the final sentiment by simply splicing the high-level semantic features of the two modalities. The experimental results show that our cross-modal block leads to a performance improvement of more than 2\% with only a 0.4M increase in the number of parameters, which indicates that efficient complementary information from both modalities can have a large impact on the final decision. Meanwhile, we observe that the output of the results by simple splicing will also be better than the early fusion method, which indicates that the early fusion method cannot adequately express the high-level semantic features of both modalities.

We further explore the validity of the internal structure of the cross-modal blocks. In the cross-modal blocks, the self-attention mechanism and the residual structure play an important role in the performance of the model. We detach the self-attention mechanism and the residual structure separately which can be seen that self-attention brings more than 2\% impact on the final result. This indicates that the audio semantic features we obtained contain redundant information and can be selected by the self-attention mechanism for feature selection to make it efficient and adaptive for inter-modal interaction. In addition, we also see that the residual structure has less impact on the final results, suggesting that the inclusion of the residual structure helps to ensure that the loss of video features is minimized during the interaction. Furthermore, we observe that the current state-of-the-art can be achieved even without the inclusion of the residual structure, which further demonstrates the efficiency of the cross-modal blocks.

More, we integrate the audio modality into the video modality as the final fusion result in our model design, denoted as $V$-\textgreater$A$. We have compared the results of integrating video modality into audio modality, denoted as $A$-\textgreater$V$. We find that there is a 1.6\% difference between them, which we attribute to the fact that the self-attention mechanism reduces the redundancy of audio features and is more effective for the results. In parallel, the rich spatio-temporal structure of the video modality also has an impact on the final result output.

\begin{table}[t]
    \fontsize{9}{11}\selectfont
    \centering
    \caption{\fontsize{9}{11}\selectfont  Ablation study on the RAVDESS dataset.
    }
    \label{tab:pool}
    \begin{tabular}{lcccc}
        \toprule
        Model & Accuracy & \#Params \\ 
        \midrule
        CFN-SR & 75.76 & 26.30M \\
        \hline
        w/o Cross-modal Blocks & 73.50 & 25.92M \\
        \hline
        V-\textgreater A Cross-modal & 74.15 & 25.67M \\
        A-\textgreater V Cross-modal & 75.76 & 26.30M \\
        \hline
        w/o self-attention & 73.86 & 26.05M \\
        w/o residual & 75.33 & 26.30M \\
        \bottomrule
    \end{tabular}%
\end{table}

\section{Conclusion}
\label{sec:concl}

In this paper, we propose a cross-modal fusion network based on self-attention and residual structure (CFN-SR) for multimodal emotion recognition. We design the cross-modal blocks, which fully considers the complementary information of different modalities and makes full use of inter-modal and intra-modal interactions to complete the transmission of semantic features. The inclusion of self-attention mechanism and residual structure can guarantee the efficiency and integrity of information interaction. We validate the effectiveness of the proposed method on the RAVDESS dataset. The experimental results show that the proposed method achieves state-of-the-art and obtains 75.76\% accuracy with 26.30M number of parameters. In the future work, we will extend the module to explore efficient interactions between multiple modalities.

\section{Acknowledgements}

This work is supported by the National Natural Science Foundation of China (No.82071171), Beijing Municipal Natural Science Foundation (No.L192026), the Science and Technology Commission of Shanghai Municipality (No.19511120601), the Scientific and Technological Innovation 2030 Major Projects (No.2018AAA0100902), the Research Project of Shanghai Science and Technology Commission (20dz2260300) and the Fundamental Research Funds for the Central Universities. We would like to thank the anonymous reviewers for their valuable suggestions.


\bibliographystyle{IEEEbib}
\bibliography{refs}

\begin{thebibliography}{10}

\bibitem{dai2021multimodal}
Wenliang Dai, Samuel Cahyawijaya, Zihan Liu, and Pascale Fung,
\newblock ``Multimodal end-to-end sparse model for emotion recognition,''
\newblock {\em arXiv preprint arXiv:2103.09666}, 2021.

\bibitem{zhang2019deep}
Yuanyuan Zhang, Zi-Rui Wang, and Jun Du,
\newblock ``Deep fusion: An attention guided factorized bilinear pooling for
  audio-video emotion recognition,''
\newblock in {\em 2019 International Joint Conference on Neural Networks
  (IJCNN)}. IEEE, 2019, pp. 1--8.

\bibitem{cao2021hierarchical}
Qi~Cao, Mixiao Hou, Bingzhi Chen, Zheng Zhang, and Guangming Lu,
\newblock ``Hierarchical network based on the fusion of static and dynamic
  features for speech emotion recognition,''
\newblock in {\em ICASSP 2021-2021 IEEE International Conference on Acoustics,
  Speech and Signal Processing (ICASSP)}. IEEE, 2021, pp. 6334--6338.

\bibitem{wu2020generalized}
Jinting Wu, Yujia Zhang, Xiaoguang Zhao, and Wenbin Gao,
\newblock ``A generalized zero-shot framework for emotion recognition from body
  gestures,''
\newblock {\em arXiv preprint arXiv:2010.06362}, 2020.

\bibitem{hernandez2021multi}
Fernanda Hern{\'a}ndez-Luquin and Hugo~Jair Escalante,
\newblock ``Multi-branch deep radial basis function networks for facial emotion
  recognition,''
\newblock {\em Neural Computing and Applications}, pp. 1--15, 2021.

\bibitem{naseem2020transformer}
Usman Naseem, Imran Razzak, Katarzyna Musial, and Muhammad Imran,
\newblock ``Transformer based deep intelligent contextual embedding for twitter
  sentiment analysis,''
\newblock {\em Future Generation Computer Systems}, vol. 113, pp. 58--69, 2020.

\bibitem{tripathi2018multi}
Samarth Tripathi, Sarthak Tripathi, and Homayoon Beigi,
\newblock ``Multi-modal emotion recognition on iemocap dataset using deep
  learning,''
\newblock {\em arXiv preprint arXiv:1804.05788}, 2018.

\bibitem{zhang2017learning}
Shiqing Zhang, Shiliang Zhang, Tiejun Huang, Wen Gao, and Qi~Tian,
\newblock ``Learning affective features with a hybrid deep model for
  audio--visual emotion recognition,''
\newblock {\em IEEE Transactions on Circuits and Systems for Video Technology},
  vol. 28, no. 10, pp. 3030--3043, 2017.

\bibitem{atmaja2020multitask}
Bagus~Tris Atmaja and Masato Akagi,
\newblock ``Multitask learning and multistage fusion for dimensional
  audiovisual emotion recognition,''
\newblock in {\em ICASSP 2020-2020 IEEE International Conference on Acoustics,
  Speech and Signal Processing (ICASSP)}. IEEE, 2020, pp. 4482--4486.

\bibitem{liu2021multimodal}
Jiaxing Liu, Sen Chen, Longbiao Wang, Zhilei Liu, Yahui Fu, Lili Guo, and
  Jianwu Dang,
\newblock ``Multimodal emotion recognition with capsule graph convolutional
  based representation fusion,''
\newblock in {\em ICASSP 2021-2021 IEEE International Conference on Acoustics,
  Speech and Signal Processing (ICASSP)}. IEEE, 2021, pp. 6339--6343.

\bibitem{sun2021multimodal}
Licai Sun, Bin Liu, Jianhua Tao, and Zheng Lian,
\newblock ``Multimodal cross-and self-attention network for speech emotion
  recognition,''
\newblock in {\em ICASSP 2021-2021 IEEE International Conference on Acoustics,
  Speech and Signal Processing (ICASSP)}. IEEE, 2021, pp. 4275--4279.

\bibitem{10.5555/3295222.3295349}
Ashish Vaswani, Noam Shazeer, Niki Parmar, Jakob Uszkoreit, Llion Jones,
  Aidan~N. Gomez, \L{}ukasz Kaiser, and Illia Polosukhin,
\newblock ``Attention is all you need,''
\newblock in {\em Proceedings of the 31st International Conference on Neural
  Information Processing Systems}, Red Hook, NY, USA, 2017, NIPS'17, p.
  6000–6010, Curran Associates Inc.

\bibitem{huang2020multimodal}
Jian Huang, Jianhua Tao, Bin Liu, Zheng Lian, and Mingyue Niu,
\newblock ``Multimodal transformer fusion for continuous emotion recognition,''
\newblock in {\em ICASSP 2020-2020 IEEE International Conference on Acoustics,
  Speech and Signal Processing (ICASSP)}. IEEE, 2020, pp. 3507--3511.

\bibitem{zhou2019exploring}
Hengshun Zhou, Debin Meng, Yuanyuan Zhang, Xiaojiang Peng, Jun Du, Kai Wang,
  and Yu~Qiao,
\newblock ``Exploring emotion features and fusion strategies for audio-video
  emotion recognition,''
\newblock in {\em 2019 International Conference on Multimodal Interaction},
  2019, pp. 562--566.

\bibitem{xie2017aggregated}
Saining Xie, Ross Girshick, Piotr Doll{\'a}r, Zhuowen Tu, and Kaiming He,
\newblock ``Aggregated residual transformations for deep neural networks,''
\newblock in {\em Proceedings of the IEEE conference on computer vision and
  pattern recognition}, 2017, pp. 1492--1500.

\bibitem{livingstone2018ryerson}
Steven~R Livingstone and Frank~A Russo,
\newblock ``The ryerson audio-visual database of emotional speech and song
  (ravdess): A dynamic, multimodal set of facial and vocal expressions in north
  american english,''
\newblock {\em PloS one}, vol. 13, no. 5, pp. e0196391, 2018.

\bibitem{neverova2015moddrop}
Natalia Neverova, Christian Wolf, Graham Taylor, and Florian Nebout,
\newblock ``Moddrop: adaptive multi-modal gesture recognition,''
\newblock {\em IEEE Transactions on Pattern Analysis and Machine Intelligence},
  vol. 38, no. 8, pp. 1692--1706, 2015.

\bibitem{wang2020speech}
Jianyou Wang, Michael Xue, Ryan Culhane, Enmao Diao, Jie Ding, and Vahid
  Tarokh,
\newblock ``Speech emotion recognition with dual-sequence lstm architecture,''
\newblock in {\em ICASSP 2020-2020 IEEE International Conference on Acoustics,
  Speech and Signal Processing (ICASSP)}. IEEE, 2020, pp. 6474--6478.

\bibitem{jin2015speech}
Qin Jin, Chengxin Li, Shizhe Chen, and Huimin Wu,
\newblock ``Speech emotion recognition with acoustic and lexical features,''
\newblock in {\em 2015 IEEE international conference on acoustics, speech and
  signal processing (ICASSP)}. IEEE, 2015, pp. 4749--4753.

\bibitem{kingma2014adam}
Diederik~P Kingma and Jimmy Ba,
\newblock ``Adam: A method for stochastic optimization,''
\newblock {\em arXiv preprint arXiv:1412.6980}, 2014.

\bibitem{ortega2019multimodal}
Juan~DS Ortega, Mohammed Senoussaoui, Eric Granger, Marco Pedersoli, Patrick
  Cardinal, and Alessandro~L Koerich,
\newblock ``Multimodal fusion with deep neural networks for audio-video emotion
  recognition,''
\newblock {\em arXiv preprint arXiv:1907.03196}, 2019.

\bibitem{fukui2016multimodal}
Akira Fukui, Dong~Huk Park, Daylen Yang, Anna Rohrbach, Trevor Darrell, and
  Marcus Rohrbach,
\newblock ``Multimodal compact bilinear pooling for visual question answering
  and visual grounding,''
\newblock {\em arXiv preprint arXiv:1606.01847}, 2016.

\bibitem{liu2018learn}
Kuan Liu, Yanen Li, Ning Xu, and Prem Natarajan,
\newblock ``Learn to combine modalities in multimodal deep learning,''
\newblock {\em arXiv preprint arXiv:1805.11730}, 2018.

\bibitem{joze2020mmtm}
Hamid Reza~Vaezi Joze, Amirreza Shaban, Michael~L Iuzzolino, and Kazuhito
  Koishida,
\newblock ``Mmtm: Multimodal transfer module for cnn fusion,''
\newblock in {\em Proceedings of the IEEE/CVF Conference on Computer Vision and
  Pattern Recognition}, 2020, pp. 13289--13299.

\bibitem{su2020msaf}
Lang Su, Chuqing Hu, Guofa Li, and Dongpu Cao,
\newblock ``Msaf: Multimodal split attention fusion,''
\newblock {\em arXiv preprint arXiv:2012.07175}, 2020.

\bibitem{verbitskiy2021eranns}
Sergey Verbitskiy and Viacheslav Vyshegorodtsev,
\newblock ``Eranns: Efficient residual audio neural networks for audio pattern
  recognition,''
\newblock {\em arXiv preprint arXiv:2106.01621}, 2021.

\end{thebibliography}

\end{document}